\documentclass[review]{arxiv}

\usepackage[nocompress]{cite}
\usepackage{array}
\usepackage[numbers]{natbib}
\usepackage{amssymb,amsmath,subcaption}
\usepackage{graphicx}\graphicspath{{figs/}{eps/}}
\usepackage{psfrag}
\usepackage{ulem}
\usepackage{algorithmic}
\usepackage{algorithm}
\usepackage{varioref}
\usepackage{booktabs}
\usepackage{tabularx}
\usepackage[para,online,flushleft]{threeparttable}

\newcommand{\mat}[1]{\left[\begin{array}{#1}}

\usepackage{url}
\begin{document}
\begin{frontmatter}

\title{Learning a Fuzzy Hyperplane Fat Margin Classifier with Minimum VC dimension}

\author[label1]{Jayadeva\corref{cor1}}
\ead{jayadeva@ee.iitd.ac.in}
\author[label2]{Sanjit S. Batra}
\author[label1]{Siddarth Sabharwal}
\cortext[cor1]{Corresponding Author}
\address[label1]{Department of Electrical Engineering, Indian Institute of Technology, Delhi, Hauz Khas, New Delhi - 110016, INDIA.}
\address[label2]{Department of Computer Science, Indian Institute of Technology, Delhi, Hauz Khas, New Delhi - 110016, INDIA.}

\begin{abstract}
The Vapnik-Chervonenkis (VC) dimension measures the complexity of a learning machine, and a low VC dimension leads to good generalization. The recently proposed Minimal Complexity Machine (MCM) learns a hyperplane classifier by minimizing an exact bound on the VC dimension. This paper extends the MCM classifier to the fuzzy domain. The use of a fuzzy membership is known to reduce the effect of outliers, and to reduce the effect of noise on learning. Experimental results show, that on a number of benchmark datasets, the the fuzzy MCM classifier outperforms SVMs and the conventional MCM in terms of generalization, and that the fuzzy MCM uses fewer support vectors. On several benchmark datasets, the fuzzy MCM classifier yields excellent test set accuracies while using one-tenth the number of support vectors used by SVMs.
\end{abstract}


\end{frontmatter}



%

\section{Introduction} \label{intro}
%
%

%
%
%
%

Support vector machines are amongst the most widely used machine learning techniques today. The most commonly used variants are the maximum margin $L_1$ norm SVM \citep{L1svm}, and the least squares SVM (LSSVM) \citep{suykens1999least}, both of which require the solution of a quadratic programming problem. The proximal SVM \citep{fung2001proximal} is also similar in spirit to the LSSVM. SVMs were motivated by the celebrated work of Vapnik and his colleagues on generalization, and the complexity of learning. The capacity of a learning machine may be measured by its VC dimension, and a small VC dimension leads to good generalization and low error rates on test data.

However, according to Burges \citep{burges1998}, SVMs can have a very large VC dimension, and that ``at present there exists no theory which shows that good generalization performance is guaranteed for SVMs''. In recent work \citep{mcmneucom}, we have shown how to learn a bounded margin hyperplane classifier, termed as the Minimal Complexity Machine (MCM) by minimizing an exact bound on its VC dimension. Experimental results on many benchmark datasets confirm that in comparison to SVMs, the MCM generalizes well while using significantly fewer support vectors, often lower by a factor between 10 and 50.

Classically, each training sample in a binary classification setting is treated equally and is associated with a unique class. However, in reality, some training samples may be corrupted by noise; this could be noise in the sample's location or in its label. Such samples may be thought of as not lying entirely in one class, but belonging to both classes to a certain degree \citep{jiang2006fuzzy}. It is well known that SVMs are very sensitive to outliers \citep{song2002robust, lin2002fuzzy, guyon1996discovering}. Fuzzy support vector machines (FSVM) \citep{lin2002fuzzy} were proposed to address this problem. In FSVMs, each sample is assigned a fuzzy membership which indicates the extent to which belongs to any one class. The membership also determines the importance of the sample in determining the separating hyperplane. Consequently, the measurement of the empirical error in a fuzzy setting does not treat all samples equally. Discounting errors on outlier samples can allow hyperplanes with larger margins to be learnt, and can also obviate the effect of noise to a considerable degree.

This paper extends the MCM into the fuzzy domain, by attempting to learn a gap tolerant, or fat margin fuzzy classifier with low VC dimension. The fuzzy MCM objective function consists of two terms. The first term is related to the VC dimension of the classifier, and minimization of this term yields a classifier with good generalization properties. The second term is a weighted sum of misclassification errors over the training samples; the weights are dependent on the fuzzy memberships of the samples, and samples that are outliers contribute less to the overall error measure. The fuzzy MCM optimization problem thus tries to find a hyperplane with a small VC dimension, that minimizes the fuzzy weighted empirical error over training data samples. The use of fuzzy memberships allows importance to be attached to individual samples, and hence helps improve generalization by not assigning equal importance to the misclassification error contributions of different samples; this reduces the effect of outliers. The fuzzy Minimal Complexity Machine, as the proposed approach is termed, dramatically outperforms conventional SVMs in terms of support vectors used, while yielding better test set accuracy. The effect of the approach to minimizing VC dimension may be guaged from the fact that on several datasets, the number of support vectors is more than fifteen times smaller than those used by SVMs. As we show in the sequel, an interesting example is that of the 'haberman' dataset from the UCI machine learning repository \citep{uciml}, that has 306 samples. A fuzzy MCM classifier learnt using 80\% of the dataset yields a classifier that can be written as a closed form expression involving only 4 support vectors. In comparison, a SVM classifier uses about 73 support vectors. \\

The rest of the paper is organized as follows. Section \ref{linmcm} briefly describes the MCM classifier, for the sake of completeness. Section \ref{lmfuzzy} shows how to extend the approach to learn a linear fuzzy MCM classifier, and section \ref{fkmcm} then extends this work to the kernel case. Section \ref{experimental} is devoted to a discussion of results obtained on selected benchmark datasets. Section \ref{conclusion} contains concluding remarks.

\section{The Linear Minimal Complexity Machine Classifier} \label{linmcm}
The motivation for the MCM originates from some outstanding work on generalization \citep{shawe1996framework, shawetaylor98, vapnik98, scholkopf2002learning}.


Consider a binary classification dataset with $n$-dimensional samples $x^i, i = 1, 2, ..., M$, where each sample is associated with a label $y_i \in \{+1, -1\}$. Vapnik \citep{vapnik98} showed that the VC dimension $\gamma$ for fat margin hyperplane classifiers with margin $d \geq d_{min}$ satisfies
\begin{equation}\label{eqnh}
\gamma \leq 1 + \operatorname{Min}(\frac{R^2}{d^2_{min}}, n)
\end{equation}
where $R$ denotes the radius of the smallest sphere enclosing all the training samples. Burges, in \citep{burges1998}, stated that \textit{``the above arguments strongly suggest that algorithms that minimize $\frac{R^2}{d^2}$ can be expected to give better generalization performance. Further evidence for this is found in the following theorem of (Vapnik, 1998), which we quote without proof''}.\\

Following this line of argument leads us to the formulations for a hyperplane classifier with minimum VC dimension; we term the same as the MCM classifier. We now summarize the MCM classifier formulation for the sake of completeness. Details may be found in \citep{mcmneucom}.

Consider the case of a linearly separable dataset. By definition, there exists a hyperplane that can classify these points with zero error. Let the separating hyperplane be given by
\begin{equation}
 u^Tx + v = 0.
\end{equation}

Let us denote
\begin{gather}
 h = \frac{\operatorname*{Max}_{i = 1, 2, ..., M} \;y_i(u^T x^i + v)}{\operatorname*{Min}_{i = 1, 2, ..., M} \;y_i(u^T x^i + v)}.
\end{gather}
In \citep{mcmneucom}, we show that there exist constants $\alpha, \beta > 0$, $\alpha, \beta \in \Re$ such that
\begin{equation}\label{exactbound}
 \alpha h^2 \leq \gamma \leq \beta h^2,
\end{equation}
or, in other words, $h^2$ constitutes a tight or exact ($\theta$) bound on the VC dimension $\gamma$. An exact bound implies that $h^2$ and $\gamma$ are close to each other.\\

\begin{figure}[hbtp]
        \centering	
                \includegraphics[scale=0.5]{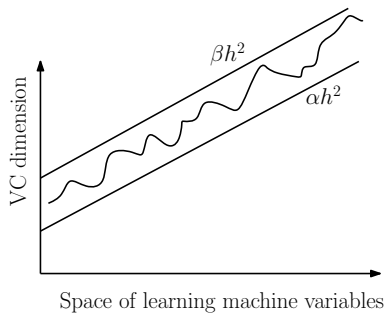}
                \caption{Illustration of the notion of an exact bound on the VC dimension. Even though the VC dimension $\gamma$ may have a complicated dependence on the variables defining the learning machine, the VC dimension $\gamma$ is bounded by multiples of $h^2$ from both above and below. The exact bound $h^2$ is thus always ``close'' to the VC dimension, and minimizing $h^2$ with respect to the variables defining the learning machine allows us to find one that has a small VC dmension. }\label{fig1}
\end{figure}

Figure \ref{fig1} illustrates this notion. It is known that the number of degrees of freedom in a learning machine is related to the VC dimension, but the connection is tenuous and usually abstruse. Even though the VC dimension $\gamma$ may have a complicated dependence on the variables defining the learning machine, the VC dimension $\gamma$ is bounded by multiples of $h^2$ from both above and below. The exact bound $h^2$ is thus always ``close'' to the VC dimension, and minimizing $h^2$ with respect to the variables defining the learning machine allows us to find one that has a small VC dmension. The use of a continuous and differentiable exact bound on the VC dimension allows us to find a learning machine with small VC dimension; this may be achieved by minimizing $h$ over the space of variables defining the separating hyperplane. In the case of a hyperplane classifier, the only variables are $u$ and $v$, and a hyperplane classifier with a small VC dimension is obtained by minimizing $h^2$ with respect to these variables.
\\

The MCM classifier solves an optimization problem, that tries to minimize the machine capacity, while classifying all training points of the linearly separable dataset correctly. This problem is given by
\begin{equation}\label{minh1}
\operatorname*{Minimize  }_{u, v} \; h ~=~ \frac{\operatorname*{Max}_{i = 1, 2, ..., M} \; y_i(u^T x^i + v)}{\operatorname*{Min}_{i = 1, 2, ..., M} \; y_i(u^T x^i + v)},
\end{equation}
that attempts to minimize $h$ instead of $h^2$, the square function $(\cdot)^2$ being a monotonically increasing one.

This optimization problem is both quasiconvex and pseudoconvex. In \citep{mcmneucom}, we further show that the optimization problem (\ref{minh1}) may be reduced to the linear programming problem
\begin{gather}
\operatorname*{Min}_{w, b, h} ~~h \label{objm4}\\
h \geq y_i \cdot [{w^T x^i + b}], ~i = 1, 2, ..., M \label{consm41}\\
y_i \cdot [{w^T x^i + b}] \geq 1, ~i = 1, 2, ..., M, \label{consm42}
\end{gather}
where $w \in \Re^n$, and $b, h \in \Re$. We refer to the problem (\ref{objm4}) - (\ref{consm42}) as the hard margin Linear Minimum Complexity Machine (Linear MCM). \\

In practice, the datasets may not be linearly separable. In such a case, we seek a classifier with a minimal VC dimension that has a small mis-classification error on the training samples. Such a hyperplane may be found by solving the soft margin MCM formulation, that is given by
\begin{gather}
\operatorname*{Min}_{w, b, h, q} ~~h + C \cdot \sum_{i = 1}^M q_i \label{objm4b}\\
h \geq y_i \cdot [{w^T x^i + b}] + q_i, ~i = 1, 2, ..., M \label{consm41b}\\
y_i \cdot [{w^T x^i + b}] + q_i \geq 1, ~i = 1, 2, ..., M, \label{consm42b}\\
q_i \geq 0, ~i = 1, 2, ..., M. \label{consm42c}
\end{gather}

Once $w$ and $b$ have been determined by solving (\ref{objm4b})-(\ref{consm42c}), the class of a test sample $x$ may be determined from the sign of the discriminant function
\begin{equation}\label{testresult}
 f(x) = w^T x + b
\end{equation}

\section{The Fuzzy Minimal Complexity Machine Classifier}\label{lmfuzzy}

In the linear soft margin MCM formulation (\ref{objm4b})-(\ref{consm42c}), the error variables $q_i, i = 1, 2, ..., M$ measure the mis-classification error on the respective data samples, and the second term of the objective function in (\ref{objm4b}) is a weighted sum of all the mis-classification errors. In this case, the hyper-parameter $C$ equally weights all variables $q_i$; this effectively means that errors made on all samples are equally important. In reality, noise tends to corrupt training samples, and robust learning requires us to ignore outliers, by assigning reduced importance to samples on which one has less confidence. 

Some samples may not be representative of a class. For example, a person showing some symptoms of a disease may have characteristics that overlap with both healthy subjects as well as unhealthy ones. Therefore, the membership of the class to which a sample belongs tends to be fuzzy, with a fuzzy membership (a value between 0 and 1) indicating the extent to which the sample may be said to belong to one class or the other. Samples with a higher membership value can be thought of as more representative of that class, while those with a smaller membership value should be given less importance when building a classifier.

\begin{figure}[hbtp]
        \centering	
                \includegraphics[scale=0.5]{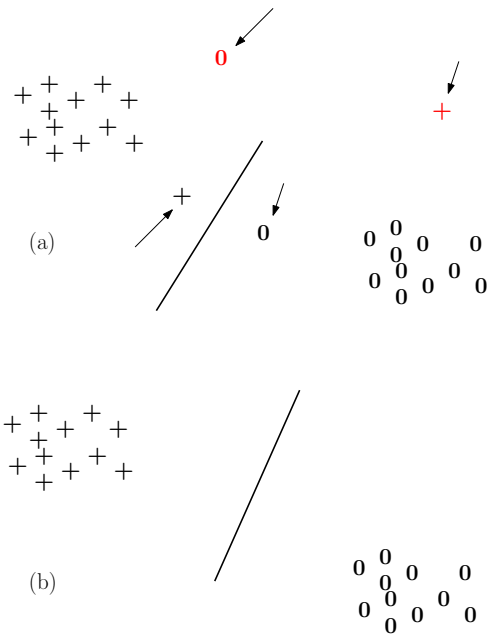}
                \caption{Discounting classification errors on outliers may allow a classifier with a smaller VC dimension to be learnt. Outliers that contribute to large classification errors may often not be representative of the class that labels indicate. Discounting classification errors on such samples allows for more robust learning. The use of fuzzy memberships provides for a natural way to measure how important it is to correctly classify a given sample.}
\label{outlierfig}
\end{figure}

Consider the samples in Fig. \ref{outlierfig}. Four outlier samples have been highlighted by arrows. The hyperplanes found before and after discounting outlier samples have been shown in (a) and (b), respectively. Two of the outliers are in black, and discounting them would allow us to obtain a hyperplane with a smaller VC dimension. Two of the outliers are marked in red, and these make the data set linearly non-separable. Discounting classification errors on these red coloured samples would allow for a more robust classifier to be learnt. 

Lin and Wang proposed fuzzy SVMs in \citep{lin2002fuzzy}, wherein they suggested that each sample be associated with a fuzzy membership $s_i$. This membership value determines how important it is to classify a data sample correctly; samples with lower values of the membership function are less representative of the class to which they have been assigned, and can therefore be mis-classified without incurring the same penalty. In the example of Fig. \ref{outlierfig}, outlier samples would have a small membership value; the optimization problem being solved factors in these membership values, thus allowing a more robust classifier to be learnt.

The fuzzy MCM classifier aims to learn a hyperplane classifier that has a small VC dimension, and that also minimizes a weighted measure of the classification error on training samples. The linear fuzzy MCM (FMCM) classifier does this by solving the following optimization problem.

\begin{gather}
\operatorname*{Min}_{w, b, h, q} ~~h + C \cdot \sum_{i = 1}^M s_i q_i \label{objf1}\\
h \geq y_i \cdot [{w^T x^i + b}] + q_i, ~i = 1, 2, ..., M \label{consf1}\\
y_i \cdot [{w^T x^i + b}] + q_i \geq 1, ~i = 1, 2, ..., M, \label{consf2}\\
q_i \geq 0 \label{consf3}
\end{gather}

Here, the fuzzy membership $s_i$ for the $i-th$ sample is used to determine the importance of the sample in terms of its possible mis-classification. This implies that samples with a small value of the fuzzy membership, such as outliers, can be ignored or accorded less importance when learning the classifier. This makes the classifier less sensitive to outliers, leading to more robust learning. In the example of Fig. \ref{outlierfig}, the values of $s_i$ for the outlier samples are small. This implies that the objective function (\ref{objf1}) discounts the errors caused when learning such samples, because of the small values of the weights $s_i$.

In the following section, we show how the linear fuzzy MCM can be extended to the kernel case.

\section{The Fuzzy Kernel MCM}\label{fkmcm}

We consider a map $\phi(x)$ that maps the input samples from $\Re^n$ to $\Re^r$, where $r > n$. The separating hyperplane in the image space is  given by

\begin{equation}
 u^T \phi(x) + v = 0.
\end{equation}

Following (\ref{objf1}) - (\ref{consf3}), the optimization problem for the fuzzy kernel MCM may be shown to be

\begin{gather}
\operatorname*{Min}_{w, b, h, q} \; h + C \cdot \sum_{i = 1}^M s_i q_i \label{objk6}\\
h \geq y_i \cdot [{w^T \phi(x^i) + b}] + q_i, ~i = 1, 2, ..., M \\
y_i \cdot [{w^T \phi(x^i) + b}] + q_i \geq 1, ~i = 1, 2, ..., M \label{consk61} \\
q_i \geq 0, ~i = 1, 2, ..., M.
\end{gather}

The image vectors $\phi(x^i), i = 1, 2, ..., M$ form an overcomplete basis in the empirical feature space, in which $w$ also lies. Hence, we can write
\begin{equation}\label{weqsumlambda}
 w = \sum_{j = 1}^M \lambda_j \phi(x^j).
\end{equation}
Note that in (\ref{weqsumlambda}), the $\phi(x^j)$'s for which the corresponding $\lambda_j$'s are non-zero may be termed as support vectors.

Therefore,
\begin{gather}\label{simpliphi}
 w^T \phi(x^i) + b = \sum_{j = 1}^M \lambda_j \phi(x^j)^T\phi(x^i) + b = \sum_{j = 1}^M \lambda_j K(x^i, x^j) + b,
\end{gather}

where $K(p, q)$ denotes the Kernel function with input vectors $p$ and $q$, and is defined as
\begin{equation}
 K(p, q) = \phi(p)^T \phi(q).
\end{equation}\label{kernel}

Substituting from (\ref{simpliphi}) into (\ref{objk6}) - (\ref{consk61}), we obtain the following optimization problem.
\begin{gather}
\operatorname*{Min}_{w, b, h, q} \; h + C \cdot \sum_{i = 1}^M s_i q_i \label{objk7}\\
h \geq y_i \cdot [\sum_{j = 1}^M \lambda_j K(x^i, x^j) + b] + q_i, ~i = 1, 2, ..., M\\
y_i \cdot [\sum_{j = 1}^M \lambda_j K(x^i, x^j) + b] + q_i \geq 1, ~i = 1, 2, ..., M \\ \label{consk71}
q_i \geq 0, ~i = 1, 2, ..., M.
\end{gather}

Once the variables $\lambda_j, j = 1, 2, ..., M$ and $b$ are obtained, the class of a test point $x$ can be determined by evaluating the sign of
\begin{equation}
f(x) ~=~ w^T \phi(x) + b ~=~ \sum_{j = 1}^M \lambda_j K(x, x^j) + b.
\end{equation}\label{testval}

Results on benchmark datasets indicate that the use of fuzzy memberships in the FMCM can reduce the number of support vectors and also lead to improved accuracies on test data. In the sequel, we present results on the linear and kernel versions of the fuzzy MCM.

\section{Experimental results}\label{experimental}
\begin{figure}[hbtp]
        \centering	
                \includegraphics[scale=0.5]{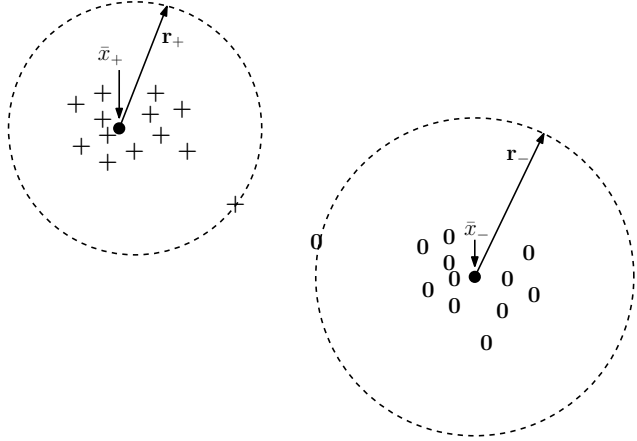}
                \caption{The figure illustrates the computation of fuzzy memberships. The fuzzy membership of a sample depends on its distance from its class centre, as well as the radius of the corresponding cluster.}
\label{fmem}
\end{figure}

The FMCM was coded in MATLAB. The code is available on request from the author. Fuzzy membership values were computed by using the approach outlined in \citep{lin2002fuzzy}. In this case, the membership value $s_i$ of the $i$-th sample is a function of its distance from its class centre. Lin and Wang suggested the formula
\begin{equation}
s_i = \begin{cases} 
       1-\frac{||\overline{x}_+ ~-~ x_i||}{r_+ ~+~ \delta}, & \mbox{if } y_i = 1 ~ \text{, i.e., the sample belongs to class 1}\\
       1-\frac{||\overline{x}_- ~-~ x_i||}{r_- ~+~ \delta}, &  \mbox{if } y_i = -1 ~ \text{, i.e., the sample belongs to class -1}
      \end{cases}
\end{equation}
Here, $r_+$ and $r_-$ are the radii of the two classes, and $\overline{x}_+$ and $\overline{x}_-$ are the respective class centres. The scalar $\delta$ is a small number used to ensure that $s_i$ does not become zero. Figure \ref{fmem} illustrates the computation of the fuzzy membership.

In order to evaluate the FMCM, we chose a number of benchmark datasets from the UCI machine learning repository \citep{uciml}. Table \ref{tabledatasets} summarizes information about the number of samples and features of each dataset.
\begin{table}[htbp]
  \centering
\caption{Characteristics of the Benchmark Datasets used}
     \begin{tabular}{|c|c|}
    \toprule
    dataset & Size (samples $\times$ features) \\
    \midrule
    fertility diagnosis & 100 $\times$ 9\\
    promoters  & 106 $\times$ 57\\
    echocardiogram  & 132 $\times$ 12\\
    hepatitis  & 155 $\times$ 19\\
    plrx  & 182 $\times$ 12\\
    heartstatlog  & 270 $\times$ 13\\
    horsecolic  & 300 $\times$ 27\\
    haberman  & 306 $\times$ 3\\
    australian  & 690 $\times$ 14\\
    crx   & 690 $\times$ 15\\
    transfusion  & 748 $\times$ 5\\
    \bottomrule
    \end{tabular}%
  \label{tabledatasets}%
\end{table}%

Table \ref{table1} summarizes five fold cross validation results of the fuzzy linear MCM on a number of datasets taken from the UCI machine learning repository. Accuracies refer to the test sets, and are indicated as mean $\pm$ standard deviation, computed using a standard five fold cross validation methodology. The table compares the linear MCM with LIBSVM using a linear kernel. The values of $C$ were determined for the FMCM by performing a grid search.
\begin{table}[htbp]
  \centering
  \caption{Linear Fuzzy MCM: Test Set Accuracies}
\begin{tabular}{ | c | c | c | }
\hline
	datasets & Linear Fuzzy MCM & Linear Fuzzy SVM \\ \hline
	haberman & 74.47 $\pm$ 3.58 & 73.87 $\pm$ 3.06 \\ \hline
	transfusion & 76.19 $\pm$ 4.12 & 76.32 $\pm$ 4.12 \\ \hline
	echocardiogram & 88.63 $\pm$ 2.44 & 84.84 $\pm$ 5.40 \\ \hline
	plrx & 71.83 $\pm$ 7.49 & 71.42 $\pm$ 7.37 \\ \hline
	crx & 70.00 $\pm$ 3.13 & 68.55 $\pm$ 2.45 \\ \hline
	horsecolic & 81.00 $\pm$ 4.03 & 80.00 $\pm$ 4.35 \\ \hline
	australian & 85.81 $\pm$ 2.01 & 85.36 $\pm$ 1.55 \\ \hline
	fertility diagnosis & 88.00 $\pm$ 9.27 & 86.00 $\pm$ 9.01 \\ \hline
	hepatitis & 67.09 $\pm$ 5.55 & 60.64 $\pm$ 7.19 \\ \hline
	pima indian diabetes & 77.33 $\pm$ 5.54 & 74.84 $\pm$ 6.62 \\ \hline
	promoters & 74.08 $\pm$ 10.88 & 69.83 $\pm$ 11.52 \\ \hline
	mammographic masses & 83.48 $\pm$ 5.13 & 79.73 $\pm$ 5.45 \\ \hline
	voting & 94.94 $\pm$ 0.92 & 94.71 $\pm$ 1.17 \\ \hline
	heart statlog & 85.18 $\pm$ 2.62 & 83.70 $\pm$ 2.72 \\ \hline
	breast & 96.83 $\pm$ 1.19 & 96.48 $\pm$ 1.24 \\ \hline
	bands & 73.86 $\pm$ 4.13 & 72.55 $\pm$ 5.00 \\ \hline
\end{tabular}
  \label{table1}%
\end{table}

Table \ref{table2} summarizes five fold cross validation results of the Fuzzy kernel MCM on a number of datasets. A Gaussian kernel was used for both the FMCM and the FSVM. The width of the Gaussian kernel was chosen by using a grid search.

\begin{table}[htbp]
  \centering
\footnotesize\setlength{\tabcolsep}{2.5pt}
  \caption{Kernel Fuzzy MCM results}
  \begin{threeparttable}
   \begin{tabular}{|c|c|c|c|c|}
    \hline
\toprule
    datasets & \multicolumn{2}{c|}{Test Set Accuracy}     & \multicolumn{2}{c|}{\# Support Vectors} \\
    \midrule
&Fuzzy Kernel MCM & Fuzzy SVM &Fuzzy Kernel MCM & Fuzzy SVM\\
\midrule

	haberman & 74.82 $\pm$ 3.72 & 72.86 $\pm$ 3.20 & 7.80 $\pm$ 6.65 & 138.20 $\pm$ 2.93 \\ \hline
	transfusion & 79.27 $\pm$ 4.20 & 77.80 $\pm$ 4.05 & 22.80 $\pm$ 11.34 & 299.06 $\pm$ 10.46 \\ \hline
	echocardiogram & 88.57 $\pm$ 6.14 & 87.13 $\pm$ 6.49 & 24.80 $\pm$ 6.14 & 48.00 $\pm$ 3.29 \\ \hline
	plrx & 71.41 $\pm$ 6.75 & 71.42 $\pm$ 7.37 & 7.00 $\pm$ 5.34 & 116.20 $\pm$ 5.49 \\ \hline
	crx & 71.01 $\pm$ 1.89 & 68.84 $\pm$ 2.79 & 92.80 $\pm$ 53.95 & 404.40 $\pm$ 7.74 \\ \hline
	horsecolic & 81.00 $\pm$ 3.27 & 79.66 $\pm$ 4.64 & 35.40 $\pm$ 15.91 & 187.20 $\pm$ 2.93 \\ \hline
	australian & 85.50 $\pm$ 1.72 & 86.08 $\pm$ 1.61 & 107.60 $\pm$ 5.92 & 244.80 $\pm$ 4.12 \\ \hline
	fertility diagnosis & 91.00 $\pm$ 8.60 & 88.00 $\pm$ 9.27 & 17.70 $\pm$ 9.09 & 39.00 $\pm$ 6.23 \\ \hline
	hepatitis & 69.03 $\pm$ 8.80 & 62.57 $\pm$ 8.31 & 44.00 $\pm$ 39.94 & 104.20 $\pm$ 2.64 \\ \hline
	pima indian diabetes & 76.55 $\pm$ 3.05 & 76.81 $\pm$ 3.31 & 112.80 $\pm$ 75.17 & 355.40 $\pm$ 7.45 \\ \hline
	promoters & 79.41 $\pm$ 3.56 & 76.46 $\pm$ 5.83 & 78.40 $\pm$ 10.71 & 84.80 $\pm$ 0.40 \\ \hline
	mammographic masses & 82.01 $\pm$ 4.50 & 82.12 $\pm$ 3.92 & 61.80 $\pm$ 9.41 & 332.00 $\pm$ 14.46 \\ \hline
    \end{tabular}%
    \end{threeparttable}
  \label{table2}%
\end{table}%

A comparison with the fuzzy SVM indicates that the fuzzy MCM yields better generalization with fewer support vectors. An examination of the table indicates that the proposed approach shows a lower test set error, and also uses a smaller number of support vectors. It is also interesting to note that the fuzzy MCM outperforms the classical MCM in terms of the number of support vectors and test set accuracies. The results of the classical MCM have not been duplicated from \citep{mcmneucom} for the sake of brevity; an added reason is that a fair comparison would be between two methods that use a fuzzy methodology.

As an interesting illustration of the sparsity of the fuzzy MCM, consider a fuzzy kernel MCM classifier using a randomly chosen subset comprising 80\% samples of the 'haberman' dataset, that employs a Gaussian kernel. This classifier may be tested by the reader on any randomly chosen set of training samples. It is interesting because it uses only four support vectors and can be written down as the following closed form expression.

\begin{align}
f(x_1, x_2, x_3) = & sign\{-105.8063 ~ exp{[-10^{-4} * ((x1 - 36)^2  + (x2 - 69)^2 + x3^2)}] \notag\\
+ & ~ 90.5143  ~ exp{[ -10^{-4} ((x1 - 43)^2  + (x2 - 58)^2 + (x3 - 52)^2)}] \notag\\
+ & ~ 129.7232  ~ exp{[ -10^{-4} ((x1 - 54)^2  + (x2 - 67)^2 + (x3 - 46)^2)}] \notag\\
- & ~ 113.7966  ~ exp{[ -10^{-4} ((x1 - 62)^2  + (x2 - 58)^2 + x3^2)}]\notag\\
- & ~ 0.7661\}
\end{align}
Here, the input samples are in three dimensions, and given by $(x_1, x_2, x_3)$.

\section{Conclusion} \label{conclusion}
In this paper, we propose a way to build a fuzzy hyperplane classifier, termed as the fuzzy Minimal Complexity Machine (MCM), that learns a fuzzy cassifier with small VC dimension. The fuzzy MCM involves the solution of a linear programming problem. Experimental results show that the fuzzy MCM outperforms the fuzzy SVM in terms of test set accuracies on a number of selected benchmark datasets. At the same time, the number of support vectors is less, often by a substantial factor, often as large as 10 or more. It has not escaped our attention that the proposed approach can be extended to fuzzy least squares classifiers, as well as to tasks such as fuzzy regression and fuzzy time series prediction; in fact, a large number of variants of fuzzy SVMs can be re-examined from the perspective of the fuzzy MCM.

\bibliographystyle{IEEEtran}
\bibliography{IEEEabrv,linear-struct-min3-els-fuzc-2}

\end{document}